\documentclass{article}

\PassOptionsToPackage{square,numbers}{natbib}
\usepackage[final]{neurips_2023}

\usepackage[utf8]{inputenc} 
\usepackage[T1]{fontenc}    
\usepackage{hyperref}       
\usepackage{url}            
\usepackage{booktabs}       
\usepackage{amsfonts}       
\usepackage{nicefrac}       
\usepackage{microtype}      
\usepackage{xcolor}         
\usepackage{graphicx}
\usepackage{makecell}
\usepackage{chngcntr}

\usepackage{float}

\title{Large Scale Masked Autoencoding for Reducing Label Requirements on SAR Data}

\author{%
  Matt Allen \\
  University of Cambridge, UK\\
  \texttt{mja78@cam.ac.uk} 
  \And
  Francisco Dorr \\
  Independent, Argentina\\
  \texttt{fran.dorr@gmail.com} 
  \And
  Joseph A. Gallego-Mejia \\
  Universidad Nacional de Colombia, Colombia\\
  \texttt{jagallegom@unal.edu.co} 
  \And
  Laura Martínez-Ferrer \\
  Universitat de Val\`encia, Spain\\
  \texttt{laura.martinez-ferrer@uv.es} 
  \And
  Freddie Kalaitzis \\
  University of Oxford, UK\\
  \texttt{freddie.kalaitzis@cs.ox.ac.uk} 
  \And
  Raúl Ramos-Pollán \\
  Universidad de Antioquia, Colombia\\
  \texttt{raul.ramos@udea.edu.co}
  \And
  Anna Jungbluth \\
  European Space Agency, Climate Office, UK\\
  \texttt{anna.jungbluth@esa.int} 
}

\begin{document}

\maketitle

\begin{abstract}
  Satellite-based remote sensing is instrumental in the monitoring and mitigation of the effects of anthropogenic climate change. Large scale, high resolution data derived from these sensors can be used to inform intervention and policy decision making, but the timeliness and accuracy of these interventions is limited by use of optical data, which cannot operate at night and is affected by adverse weather conditions. Synthetic Aperture Radar (SAR) offers a robust alternative to optical data, but its associated complexities limit the scope of labelled data generation for traditional deep learning. In this work, we apply a self-supervised pretraining scheme, masked autoencoding, to SAR amplitude data covering 8.7\% of the Earth's land surface area, and tune the pretrained weights on two downstream tasks crucial to monitoring climate change - vegetation cover prediction and land cover classification. We show that the use of this pretraining scheme reduces labelling requirements for the downstream tasks by more than an order of magnitude, and that this pretraining generalises geographically, with the performance gain increasing when tuned downstream on regions outside the pretraining set. Our findings significantly advance climate change mitigation by facilitating the development of task and region-specific SAR models, allowing local communities and organizations to deploy tailored solutions for rapid, accurate monitoring of climate change effects.
\end{abstract}

\section{Introduction}
Satellite remote sensing has fundamentally changed the way we address climate change, offering large scale, high resolution data for applications such as forest mapping \cite{immitzer_first_2016}\cite{waldeland_forest_2022}, wildfire monitoring \cite{hu_sentinel-2_2021} and flood detection \cite{tarpanelli_effectiveness_2022}. Data from such tasks is crucial to address problems caused by climate change, but is restricted by the limitations of optical sensing. The inability of these sensors to operate at night, through cloud cover and without atmospheric interference mean that optical data is inappropriate for time sensitive tasks such as natural disaster management \cite{tarpanelli_effectiveness_2022}. Synthetic Aperture Radar (SAR) overcomes these limitations, providing more consistent, all-weather, day-night monitoring capabilities. These enhanced capabilities are invaluable for timely intervention in situations including extreme weather \cite{otto_attribution_2016}, natural disasters \cite{clarke_extreme_2022}, rapid ecological shifts \cite{allen_underestimation_2015}, and deforestation \cite{moffette_impact_2021}, all of which have implications for climate change.

While SAR's robust capabilities offer a promising avenue for overcoming the challenges associated with optical sensors, it comes with its own set of complexities. The technical demands associated with processing SAR data, including aspects like coherence estimation and interferogram formation, make it challenging to apply conventional machine learning techniques. Such hurdles limit the ease of generating labeled data for supervised learning, limiting SAR's effectiveness in automated analysis.

Self-supervised learning offers the advantage of learning directly from the input data without requiring ground truth labels. Methodologies such as those based on contrastive learning \cite{radford_learning_2021}, masked image modelling \cite{he_masked_2021} and knowledge distillation \cite{caron_emerging_2021} have achieved remarkable successes on RGB image data, yielding significant improvements in various tasks such as image classification, segmentation, and object detection \cite{wang_image_2022}\cite{yu_coca_2022}, reducing dependency on labelled data. Despite these advancements, the application of self-supervised learning to SAR data remains relatively unexplored \cite{wang_self-supervised_2022}. Approaches based on data fusion with other remote sensing data sources such as RGB satellite or aerial imagery exist \cite{chen_self-supervised_2022}\cite{sun_ringmo_2023}, but although these approaches can exploit the specifics of SAR data, they may not be robust to the absence of usable RGB data at night or under cloud cover. A small number of methods operating solely on SAR data exist \cite{ren_mutual_2021}\cite{wen_rotation_2021}\cite{xu_adversarial_2021}, but have not yet clearly shown the geographic or temporal generalisabilty often lacking in remote sensing models \cite{safonova_ten_2023}. Applying self-supervised learning directly to the large amounts of available unlabelled SAR data would allow practitioners to circumvent the limitations posed by the absence of reliable RGB data at night or in cloudy conditions - improving accuracy and response time in areas such as disaster management and environmental monitoring. Moreover, the use of large-scale, geographically diverse data with a model large enough to accommodate it has the potential to overcome the generalisability issues that often plague remote sensing models, presenting a robust alternative solely based on SAR data.

In this work, we take a self-supervised pretraining scheme - masked autoencoding \cite{he_masked_2021} - that has been proven effective on curated RGB imagery, and apply it to polarimetric SAR data on a large set of data covering 8.7\% of the Earth's land surface. We finetune the pretrained model on two downstream tasks - vegetation cover prediction (per-image regression) and land cover classification (semantic segmentation). We show that, in all cases, pretraining improved performance on downstream tasks. We also show that models initialized with pretrained weights still outperform their randomly initialized counterparts when using substantially fewer labels. We show that the pretrained model generalised well to regions that were not seen in the pretraining set.

\section{Methods}
\subsection{Data}
\paragraph{Split}
Our data comprises four areas of interest (AOIs) - China, the Continental United States (CONUS), Europe and South America. Of these AOIs, three comprise the pretraining set (Europe, CONUS, China). For each AOI, we divide imagery and labels into tiles of size 4480m$\times$4480m and split the resulting tiles using geographic bands into train, validation and test sets, to avoid data leakage on contiguous tiles as much as possible (Appendix~\ref{app:datasplit}). We used data from 2020 exclusively in this work to avoid the computational expense of preprocessing datasets from multiple years. Since the Earth is finite, it is feasible to pretrain a model on the entire planet, so future work should focus whether our approach also generalises temporally.

\paragraph{Input Data}\label{sec:input}
For all tasks, we derived input data from ESA Sentinel-1 Level 1 Ground Range Detected SAR (S1GRD) amplitude data, tiled from Google Earth Engine using \texttt{geetiles}\footnote{\url{https://github.com/rramosp/geetiles}} We used seasonal averages (spring, summer, autumn, winter) in two acquisition modes and their logarithmic difference (VV, VH, VV-VH) as input, totalling 12 channels. The resolution of S1GRD imagery is approximately 10m/pixel. 
\paragraph{Task Labels}
\textbf{Vegetation cover} percentage labels were obtained from the Terra MODIS Vegetation Continuous Fields product (MODISVEG), available in Google Earth Engine as MOD44B.006\footnote{\url{https://developers.google.com/earth-engine/datasets/catalog/MODIS\_006\_MOD44B}}. The resolution of the MODISVEG product is approximately 250m/pixel. We predicted the mean value of vegetation cover within each tile (percentage area covered by vegetation). \textbf{Land cover classification} labels were obtained from the ESA World Cover product (ESAWC), also from Google Earth Engine\footnote{\url{https://developers.google.com/earth-engine/datasets/catalog/ESA\_WorldCover\_v200}}. The resolution of the ESAWC dataset is approximately 10m/pixel, and spans 11 land cover classes. We report segmentation accuracy using mean intersection-over-union (mIoU). For both tasks, we evaluated downstream performance on one region within (Europe) and one outside (South America) the pretraining set. In all cases we trained one model from scratch and one with an encoder pretrained using masked autoencoding. We do not compare results directly to other work on the same datasets to avoid conflating performance differences due to the methods and architectures we chose when developing our model with those due to differences in input data type or SAR preprocessing differences.

\subsection{Models}\label{sec:models}
For pretraining we used a masked autoencoder with a ViT-B \cite{dosovitskiy_image_2021} encoder followed by a reconstruction decoder based on \cite{he_masked_2021}. We motivate our model selection by the observation that the original masked autoencoder behaves reasonably with minimal data augmentation \cite{he_masked_2021}. Selecting appropriate data augmentations for SAR data introduces additional complexity compared to RGB data - for example, rotation or flipping may introduce invariance to information specific to each polarisation of the instrument. A contrastive approach based on two different SAR modes \cite{radford_learning_2021} - for example, polarimetry with two different polarisations or polarimetry and another mode such as coherence - may similarly neglect information specific to each mode. We therefore choose to omit comparison to additional methods in this short work, although future comparison remains of interest.

We applied two modifications to the model - we use 12 channels, as described in Section~\ref{sec:input}, and reduce the patch size from 32 to 16 - the same size in pixels as per the original implementation, but smaller relative to our input image size of $448\times448$, therefore resulting in a longer sequence input to the transformer encoder. We motivate this change with the observation that distant pixels in remote sensing imagery are less likely to be correlated than distant pixels in a curated photograph. See Appendices~\ref{app:reconstructions} and~\ref{app:training} for qualitative reconstruction results and hyperparameter details.

For the MODISVEG task, we replaced the reconstruction decoder with a regression head comprising 1D convolutions, of output dimension 196, in both the sequence and hidden dimensions, followed by 3 fully connected layers of sizes \{512, 256, 128\}. We use ReLU \cite{agarap_deep_2019} activation functions between hidden layers and ELU \cite{clevert_fast_2016} before the regression output.

For the ESAWC task, we followed SETR-PUP \cite{zheng_rethinking_2021}. We increase the number of decoder layers compared to the original implementation to maintain a maximum upsampling of $2\times$ per layer.


\section{Results}
\begin{figure}[h]
    \includegraphics[width=\textwidth]{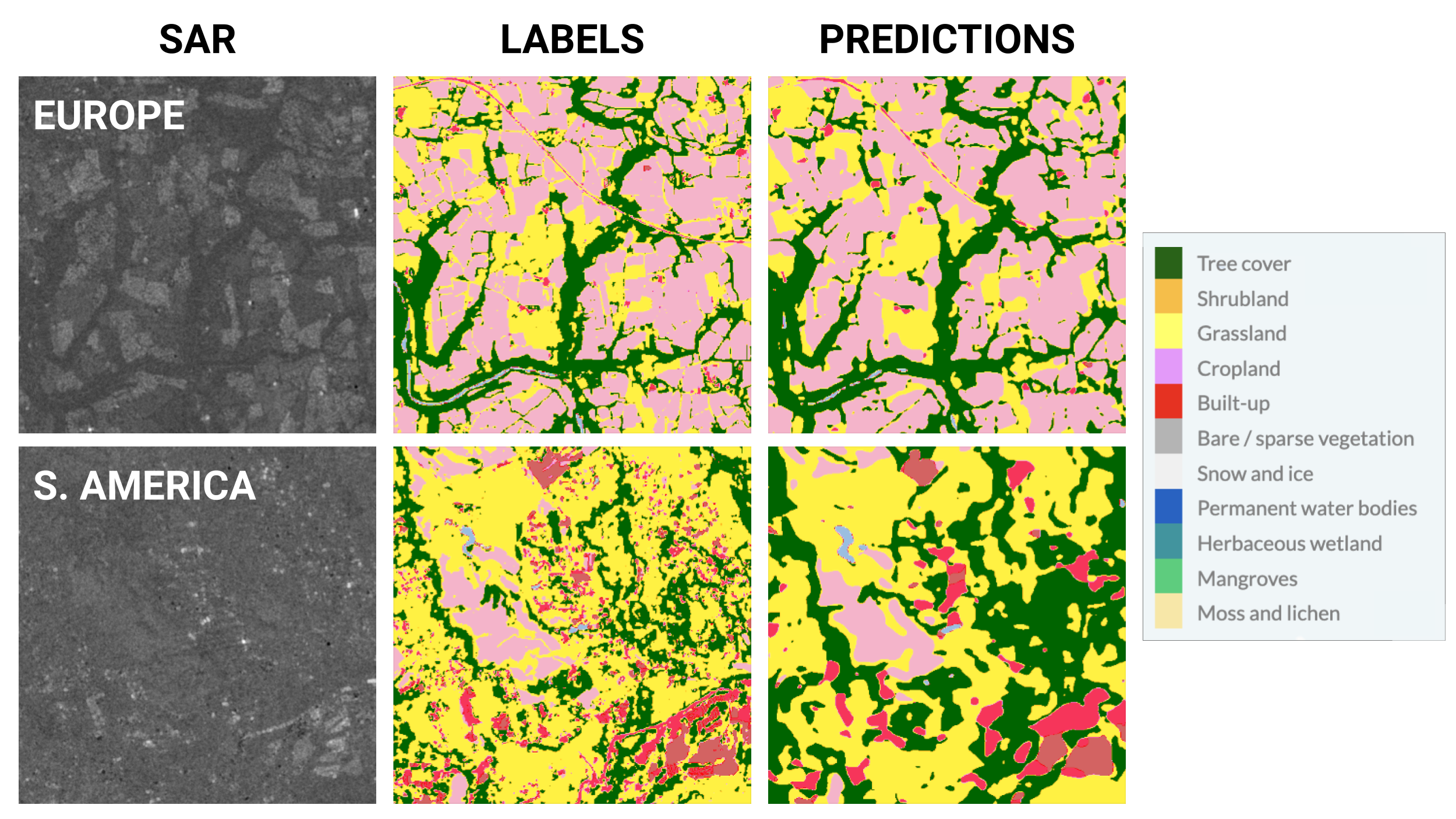}
  \caption{\textbf{Qualitative results for ESAWC land cover classification: } Land cover classification for ESAWC on data from Europe (top row) and South America (bottom row).\label{fig:esawc-fig}}
\end{figure}

\paragraph{ESAWC} Quantitative results for the ESAWC task are presented in Table~\ref{tab:esawc-table}, and qualitative results in Figure~\ref{fig:esawc-fig}. In the fully supervised cases, both with and without pretraining and in the regions within (Europe) and outside (South America) the pretraining set, the model was skillful in classifying land cover (Best mIoU Europe: $0.533$, South America: $0.426$). For a fixed number of labels, in all cases, performance was improved by pretraining the encoder using masked autoencoding. The effect of pretraining on downstream performance increased when the downstream task inputs were from outside the pretraining set - using the South American data downstream, the model using the pretrained encoder and 10\% of the labelled data (mIoU: 0.399) outperformed the randomly initialised, fully supervised model trained with 100\% of available data (mIoU: 0.393). When performing the same ablation on the European data, the pretrained model using 10\% of the downstream labels (mIoU: 0.508) did not outperform the randomly initialised, fully supervised model (mIoU: 0.522).

\begin{table}[h]
  \caption{\textbf{ESA World Cover (ESAWC) segmentation accuracy reported as mIoU}: European data is in the pretraining set, and South American data is unseen before the downstream task. Best results for each region in \textbf{bold}.}
  \label{tab:esawc-table}
  \centering
  \begin{tabular}{llllll}
    \toprule
    & & \multicolumn{2}{c}{Europe} & \multicolumn{2}{c}{South America} \\
    \cmidrule(lr){3-4} \cmidrule(lr){5-6}
    & & Scratch & Pretrained & Scratch & Pretrained \\
    \midrule
    0.1\% & & 0.345 & 0.405 & 0.214 & 0.267 \\
    1\%   & & 0.438 & 0.475 & 0.290 & 0.360 \\
    10\%  & & 0.486 & 0.508 & 0.339 & 0.399 \\
    100\% & & 0.522 & \textbf{0.533} & 0.393 & \textbf{0.426} \\
    \bottomrule
  \end{tabular}
\end{table}

  

  

\begin{figure}[h]
  \centering
  \begin{minipage}[b]{.49\textwidth}
    \includegraphics[width=\textwidth]{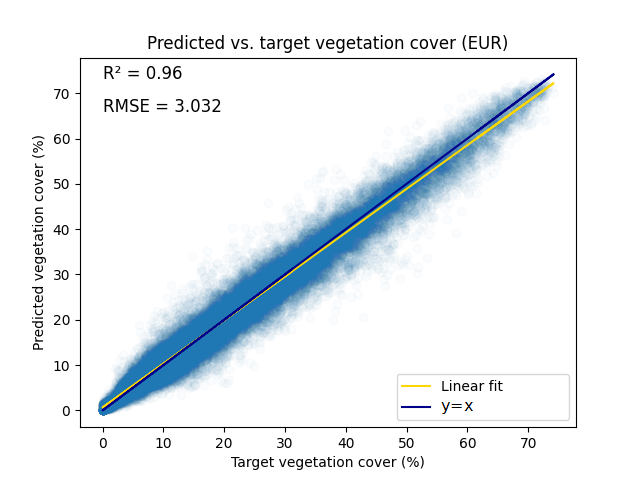}
  \end{minipage}\hspace{0.01\textwidth} 
  \begin{minipage}[b]{.49\textwidth}
    \includegraphics[width=\textwidth]{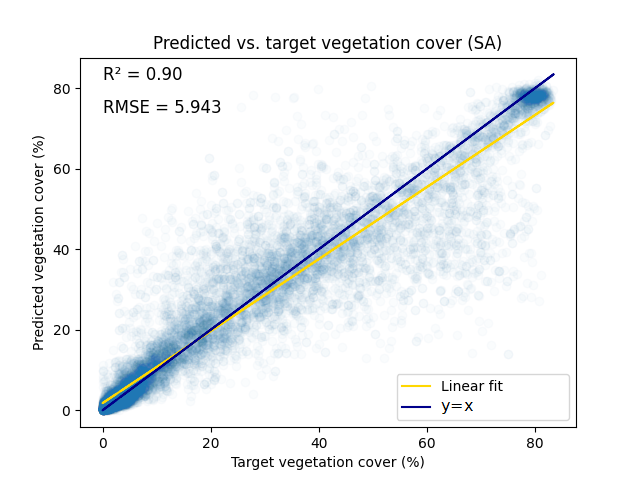}
  \end{minipage}
  
  \caption{\textbf{Correlation plots for MODISVEG prediction finetuning the pretrained model with 100\% of the labelled data: } Europe (left column) and South America (right column). Linear fits obtained by ordinary least squares (OLS). \label{fig:modisveg-fig}}
\end{figure}

\paragraph{MODISVEG} Quantitative results for the MODISVEG task are presented in Table~\ref{tab:modisveg-table}, and correlation plots in Figure~\ref{fig:modisveg-fig}. In all fully supervised cases the model was skillful in predicting mean vegetation cover percentage, and pretraining using masked autoencoding improved performance in all cases. The effect of pretraining was very strong for this task - the pretrained model needed an order of magnitude less data than the randomly initialised model to achieve the same or greater performance for all data percentages in both regions. Again, the effect of pretraining increased in the region outside of the pretraining set (South America), with the pretrained model tuned using 1\% of the task labels (RMSE 8.390 Veg \%) outperforming the fully supervised randomly initialised model (RMSE 8.883 Veg \%). For the model tuned on the European data, the pretrained model using 10\% of the task labels (RMSE 3.282 Veg \%) outperformed the fully supervised, randomly initialised model (RMSE 3.749 Veg \%), 

\begin{table}[H]
  \caption{\textbf{Terra MODIS Vegetation (MODISVEG) prediction reported as RMSE (mean vegetation cover \%):} European data is in the pretraining set, and South American data is unseen before the downstream task. Best results for each region in \textbf{bold}.}
  \label{tab:modisveg-table}
  \centering
  \begin{tabular}{llllll}
    \toprule
    & & \multicolumn{2}{c}{Europe} & \multicolumn{2}{c}{South America} \\
    \cmidrule(lr){3-4} \cmidrule(lr){5-6}
    & & Scratch & Pretrained & Scratch & Pretrained \\
    \midrule
    0.1\% & & 7.187 & 5.138 & 14.326 & 10.902 \\
    1\%   & & 5.671 & 4.082 & 12.415 & 8.390 \\
    10\%  & & 4.171 & 3.282 & 10.116 & 7.256 \\
    100\% & & 3.749 & \textbf{3.032} & 8.883 & \textbf{5.943} \\
    \bottomrule
  \end{tabular}
\end{table}

although the model tuned with 1\% of the task labels did not (RMSE 4.082 Veg \%). It is unclear if the improved label efficiency for regions outside the training set is due to geographic diversity or due to the encoder being trained on a larger combined set of pretraining and training tiles.

\section{Conclusions}
Satellite remote sensing with Synthetic Aperture Radar (SAR) offers significant advantages over optical sensors, notably the ability to operate in all-weather conditions and during both day and night. These capabilities are essential for timely responses in climate change mitigation and natural disaster management. Processing and labelling this data, however, is subject to substantially more complexity. In this context, we showed that self-supervised pretraining on SAR data using masked autoencoding dramatically reduces the label requirements for effective performance in downstream tasks. The benefits of pretraining were particularly pronounced for geographic regions not seen during pretraining. By reducing label requirements and improving geographic generalisability, our work enables the application of deep learning to SAR for all-weather, day-night monitoring - significantly improving our capability to address climate change on a near-real-time basis. This enhanced monitoring frequency is crucial during extreme weather events, natural disasters, and rapid ecological changes, allowing for more timely intervention and mitigation strategies.

\begin{ack}
This work has been enabled by Frontier Development Lab Europe (\url{https://fdleurope.org}) a public / private partnership between the European Space Agency (ESA), Trillium Technologies, the University of Oxford and leaders in commercial AI supported by Google Cloud and Nvidia, developing open science for all Humankind.  L.M-F. was supported by the European Research Council (ERC) Synergy Grant “Understanding and Modelling the Earth System with Machine Learning (USMILE)” under the Horizon 2020 research and innovation programme (Grant agreement No. 855187). M. J. A. was supported by the UKRI Centre for Doctoral Training in Application of Artificial Intelligence to the study of Environmental Risks [EP/S022961/1], and additionally by Trinity Hall, Cambridge. We are also indebted to Nicolas Longépé, Carlos López-Martínez, Fabio A. González Osorio, Samuel Bancroft, Emma Hatton, Alison Lowndes, Alistair Francis, Ioanna Bouri and the rest of reviewers during 2023 FDL-Europe sprint.  
\end{ack}

\medskip

{
\small
\bibliography{NeurIPS}
}
\newpage
\appendix
\counterwithin{figure}{section}
\counterwithin{table}{section}
\section{Supplementary Material}

\subsection{Data Split \label{app:datasplit}}
\begin{figure}[h]
  \centering
    \includegraphics[width=\textwidth]{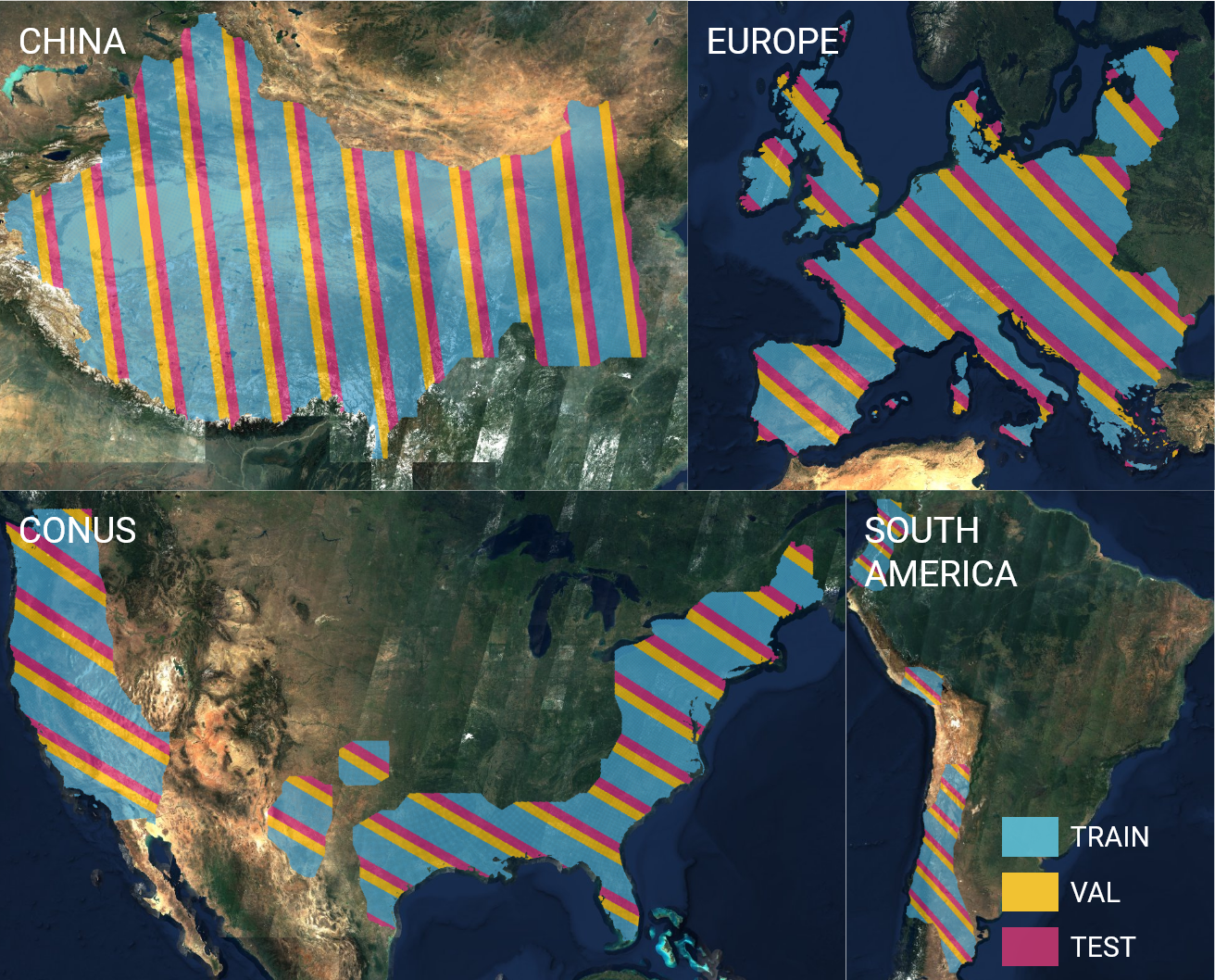}
  
  \caption{\textbf{Data split}: Geographic bands for training/validation/test sets at a ratio of 60:20:20.\label{fig:sardata-fig}}
\end{figure}

We used repeated geographic bands to define our training, validation and test sets. These bands can be seen for the four AOIs in Figure~\ref{fig:sardata-fig}. Coverage was determined by intersection with the coverage of the ARIA S1 GUNW dataset\footnote{\url{https://asf.alaska.edu/data-sets/derived-data-sets/sentinel-1-interferograms/}}, which was not used in this work. This approach minimises data leakage compared with a fully randomised split, while also reducing the train-test distribution shift that would occur when using one geographically contiguous band for each set. Data was split into the training, validation and test sets at a 60:20:20 ratio. A total of 737,050 tiles were generated, spanning an area of 1.4793$\times 10^7$km$^2$. See Table~\ref{tab:sardata-table} for a breakdown by AOI.

\begin{table}[h]
  \caption{\textbf{Statistics for S1GRD dataset tiles:}. Regions in the pretraining set are shown in \textbf{bold}. Regions used in downstream tasks are shown in \textit{italics}.}
  \label{tab:sardata-table}
  \centering
  \begin{tabular}{lllll}
    \toprule
    AOI & Total No. Tiles & Area (km$^2$) & \% Earth's & Size (GB) \\
        &                 &               & land surface  &  \\
    \midrule
    \textbf{China} & \textbf{285402} & \textbf{5.728$\mathbf{\times 10^6}$} & \textbf{3.7\%} & \textbf{1740} \\
    \textbf{Conus} & \textbf{167403} & \textbf{3.360$\mathbf{\times 10^6}$} & \textbf{2.3\%} & \textbf{1003} \\
    \textbf{\textit{Europe}} & \textbf{\textit{200489}} & \textbf{\textit{4.024$\mathbf{\times 10^6}$}} & \textbf{2.7\%} & \textbf{\textit{1228}} \\
    \textit{South America} & \textit{83756} & \textit{1.681$\times 10^6$} & 1.1\% &\textit{502} \\
    \midrule
    \textbf{Pretrain} & \textbf{653294} & \textbf{12.112$\mathbf{\times 10^6}$} & \textbf{8.7\%} & \textbf{3971} \\
    Total & 737050 & 14.793$\times 10^6$ & 9.8\% & 4473 \\
    \bottomrule
  \end{tabular}
\end{table}

\newpage
\subsection{SAR Reconstructions \label{app:reconstructions}}

\begin{figure}[h]
  \centering
  
  \begin{minipage}[b]{.2\textwidth}
    \includegraphics[width=\textwidth]{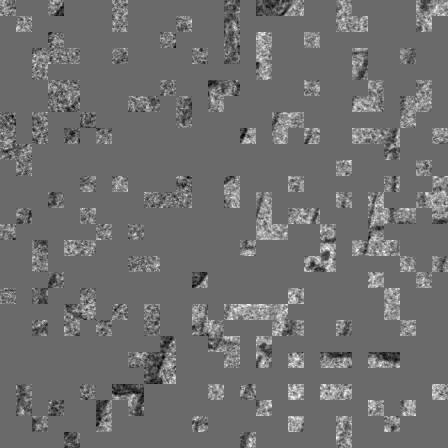}
  \end{minipage}\hspace{0.01\textwidth}
  \begin{minipage}[b]{.2\textwidth}
    \includegraphics[width=\textwidth]{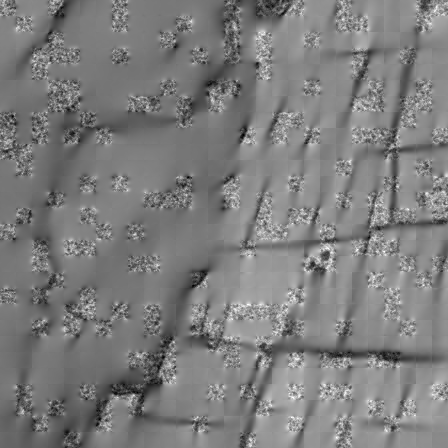}
  \end{minipage}\hspace{0.01\textwidth}
  \begin{minipage}[b]{.2\textwidth}
    \includegraphics[width=\textwidth]{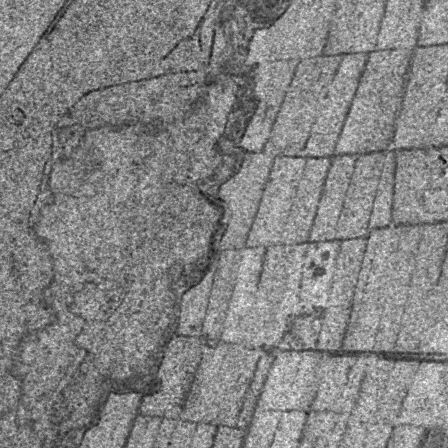}
  \end{minipage}
  
  \vspace{0.01\textwidth}
  
  \begin{minipage}[b]{.2\textwidth}
    \includegraphics[width=\textwidth]{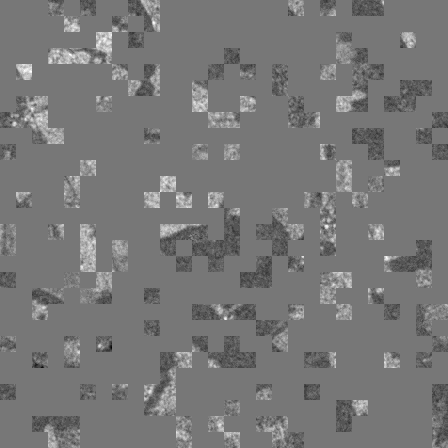}
  \end{minipage}\hspace{0.01\textwidth}
  \begin{minipage}[b]{.2\textwidth}
    \includegraphics[width=\textwidth]{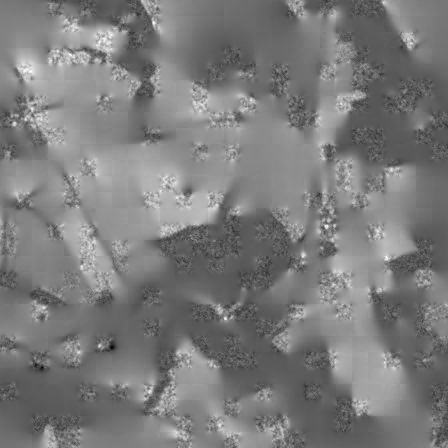}
  \end{minipage}\hspace{0.01\textwidth}
  \begin{minipage}[b]{.2\textwidth}
    \includegraphics[width=\textwidth]{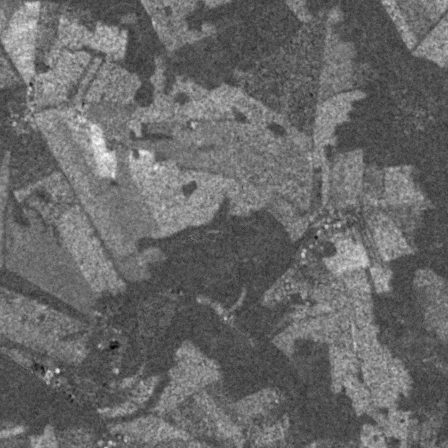}
  \end{minipage}

  \vspace{0.01\textwidth}
  
  \begin{minipage}[b]{.2\textwidth}
    \includegraphics[width=\textwidth]{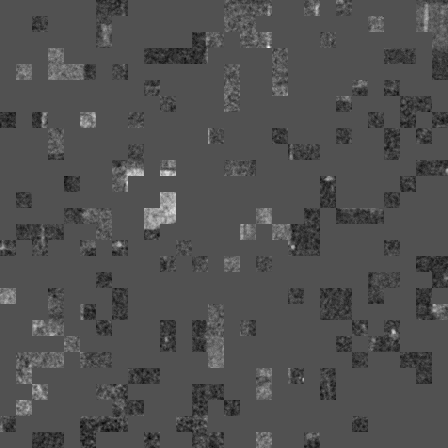}
  \end{minipage}\hspace{0.01\textwidth}
  \begin{minipage}[b]{.2\textwidth}
    \includegraphics[width=\textwidth]{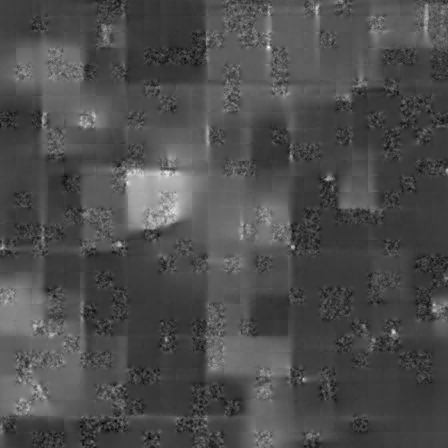}
  \end{minipage}\hspace{0.01\textwidth}
  \begin{minipage}[b]{.2\textwidth}
    \includegraphics[width=\textwidth]{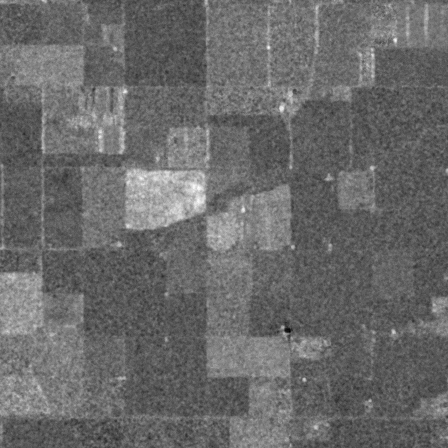}
  \end{minipage}
  
  \vspace{0.01\textwidth}
  
  \begin{minipage}[b]{.2\textwidth}
    \includegraphics[width=\textwidth]{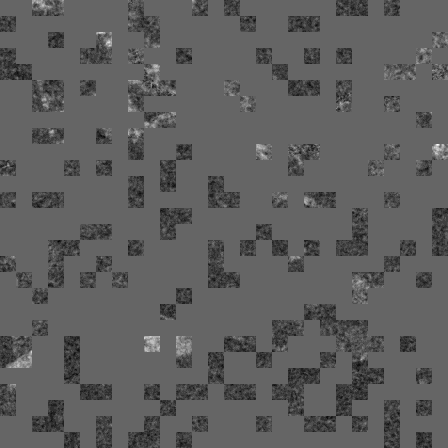}
  \end{minipage}\hspace{0.01\textwidth}
  \begin{minipage}[b]{.2\textwidth}
    \includegraphics[width=\textwidth]{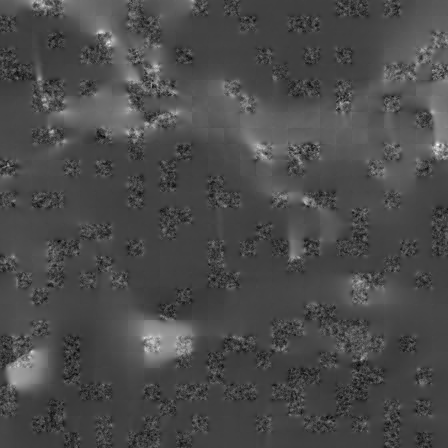}
  \end{minipage}\hspace{0.01\textwidth}
  \begin{minipage}[b]{.2\textwidth}
    \includegraphics[width=\textwidth]{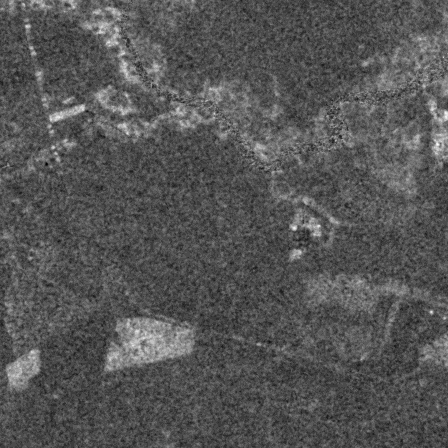}
  \end{minipage}
  
  \vspace{0.01\textwidth}
  
  \begin{minipage}[b]{.2\textwidth}
    \includegraphics[width=\textwidth]{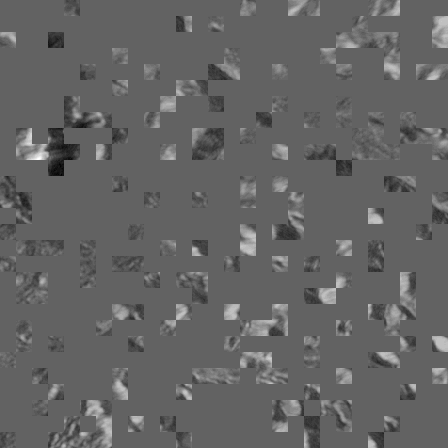}
  \end{minipage}\hspace{0.01\textwidth}
  \begin{minipage}[b]{.2\textwidth}
    \includegraphics[width=\textwidth]{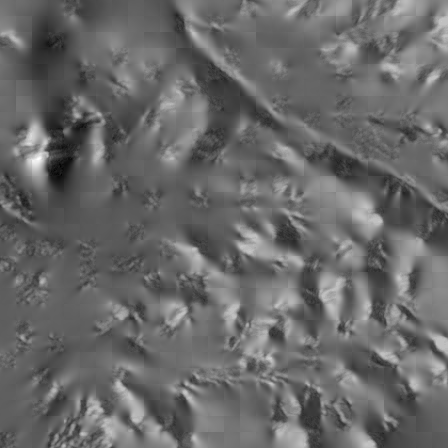}
  \end{minipage}\hspace{0.01\textwidth}
  \begin{minipage}[b]{.2\textwidth}
    \includegraphics[width=\textwidth]{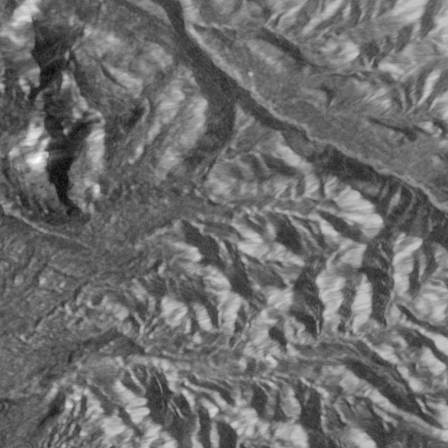}
  \end{minipage}

  \vspace{0.01\textwidth}
  
  \begin{minipage}[b]{.2\textwidth}
    \includegraphics[width=\textwidth]{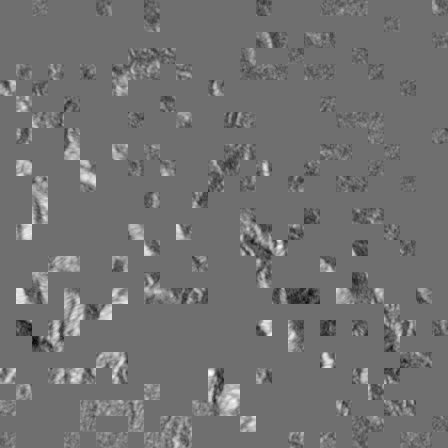}
  \end{minipage}\hspace{0.01\textwidth}
  \begin{minipage}[b]{.2\textwidth}
    \includegraphics[width=\textwidth]{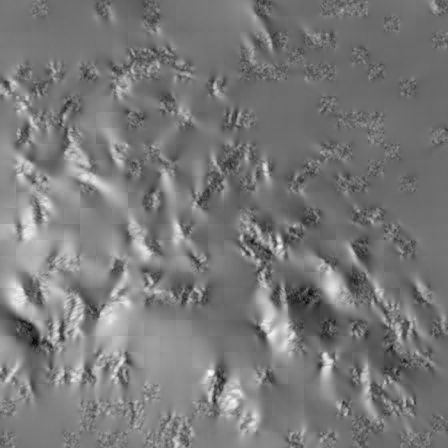}
  \end{minipage}\hspace{0.01\textwidth}
  \begin{minipage}[b]{.2\textwidth}
    \includegraphics[width=\textwidth]{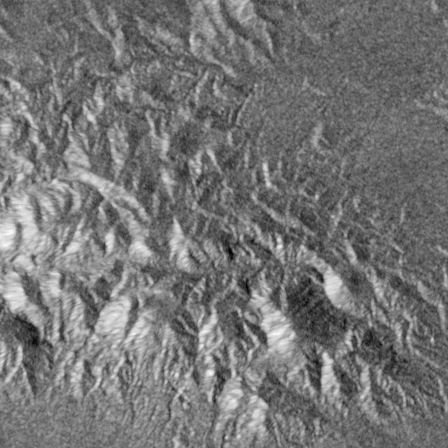}
  \end{minipage}
  
  \caption{\textbf{Masked SAR Reconstruction:} Masked autoencoder-based reconstruction of SAR amplitude imagery from the validation set. Within each row, we show the masked image (left), reconstruction (centre) and original image (right). A masking ratio of 0.75 was applied to patches of size $16\times16$ on images of size $448\times448$.}
  \label{fig:rec-fig}
\end{figure}

Reconstructions by the masked autoencoder of SAR data masked during pretraining can be seen in Figure~\ref{fig:rec-fig}. Note that the explicit aim of pretraining is to learn input features, not to obtain high reconstruction accuracy. The model largely predicts low-frequency features, as in~\cite{he_masked_2021}.

\subsection{Training Details \label{app:training}}

Hyperparameter details for MAE-based pretraining and task finetuning can be seen in Table~\ref{comparison-table}. The patch size was halved relative to the size of the image compare to~\cite{he_masked_2021}, based on the intuition that distant pixels in remote sensing imagery are less likely to be correlated than distant pixels in curated photographs. This intuition appeared to be evidenced by linear probe performance on the validation data for vegetation prediction, although we do not report the results from the linear probe as it was not used on all tasks or regions. Beyond this, the probe was not used to make further design decisions. Increasing or decreasing the learning rate by an order of magnitude did not improve convergence on the validation data. No significant hyperparameter tuning was undertaken beyond these two decisions, as the computational expense of performing an equal level of tuning for all tasks and regions was too high.

\begin{table}[h]
  \renewcommand{\arraystretch}{1.2}
  \caption{\textbf{Hyperparameter details} for MAE pretraining and downstream training on MODISVEG and ESAWC.}
  \label{comparison-table}
  \centering
  \begin{tabular}{l l l l}
    \toprule
    \textbf{} & \textbf{MAE Pretraining} & \textbf{MODISVEG} & \textbf{ESAWC} \\
    \midrule
    \makecell[l]{\textbf{Encoder} \\ (Params)} & \makecell[l]{ViT-B \cite{dosovitskiy_image_2021} \\ (88.8M)} & \makecell[l]{ViT-B \cite{dosovitskiy_image_2021} \\ (88.8M)} & \makecell[l]{ViT-B \cite{dosovitskiy_image_2021} \\ (88.8M)} \\
    \makecell[l]{\textbf{Decoder} \\ (Params)} & \makecell[l]{Reconstruction \cite{he_masked_2021} \\ (5.5M)} & \makecell[l]{$1\times1$ Conv + FC \\ (Section~\ref{sec:models}; 19.7M)} & \makecell[l]{SETR-PUP \cite{zheng_rethinking_2021} \\ (3.0M)} \\
    \textbf{Loss Function} & MSE & RMSE & Cross Entropy \\
    \textbf{Input Image Size} & $448\times448$ & $448\times448$ & $448\times448$ \\
    \textbf{Output Image Size} & $448\times448$ & Scalar & $448\times448$ \\
    \textbf{Patch Size} & $16\times16$ & $16\times16$ & $16\times16$ \\
    \textbf{Masking Type} & Random & N/A & N/A \\
    \textbf{Masking Ratio} & 0.75 & N/A & N/A \\
    \textbf{Optimiser} & AdamW & AdamW & AdamW \\
    \textbf{Learning Rate} & 1.00E-04 & 1.00E-04 & 1.00E-04 \\
    \midrule
    \textbf{No. Epochs (Pretrain)} & 75 & N/A & N/A \\
    \textbf{No. Epochs (0.1\%)} & N/A & 75 & 100 \\
    \textbf{No. Epochs (1\%)} & N/A & 50 & 75 \\
    \textbf{No. Epochs (10\%)} & N/A & 50 & 50 \\
    \textbf{No. Epochs (100\%)} & N/A & 25 & 35 \\
    \bottomrule
  \end{tabular}
\end{table}

\subsection{Training Histories \label{app:convergence}}

\begin{figure}[h]
    \includegraphics[width=\textwidth]{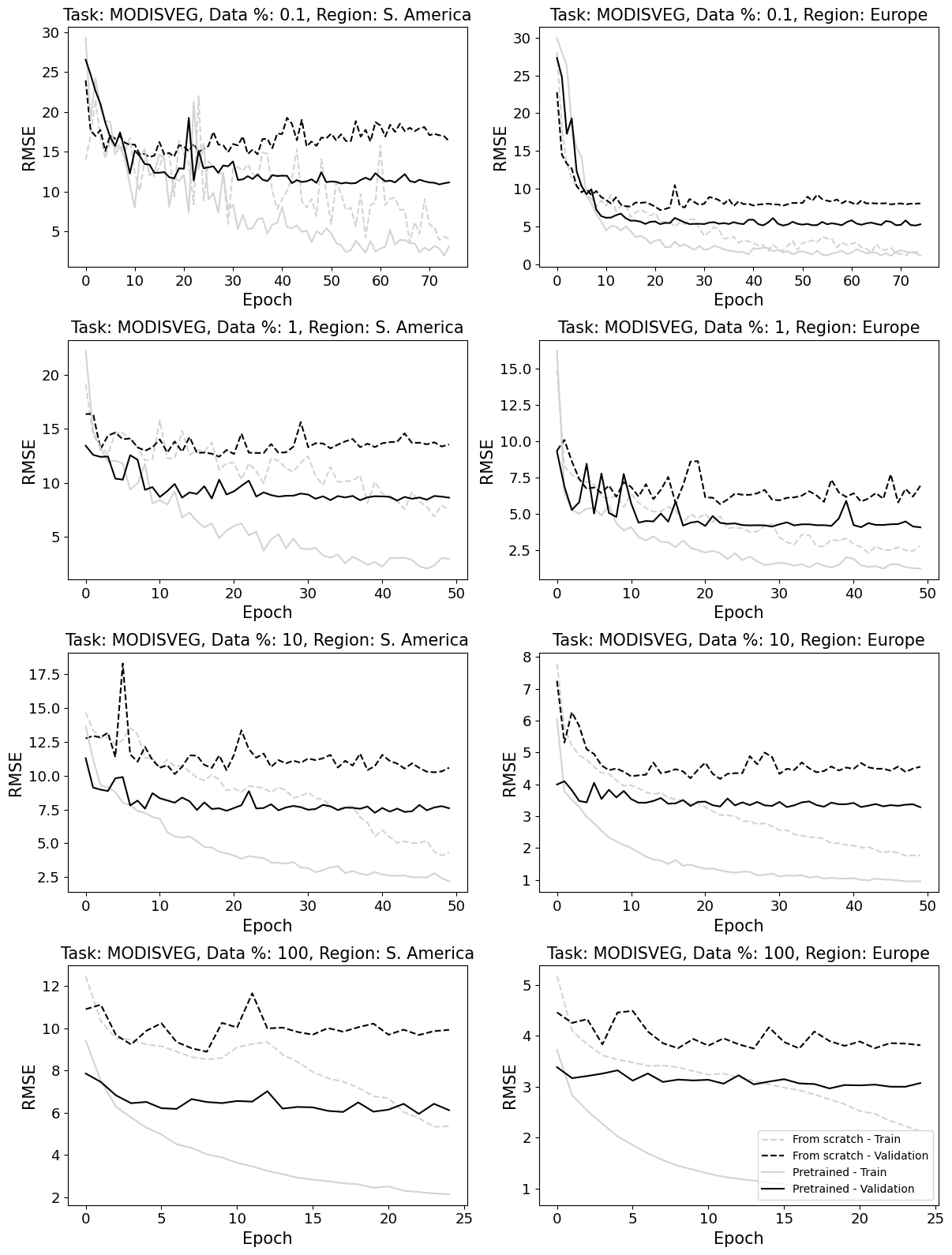}
  \caption{Training histories for Terra MODIS Vegetation prediction. \textbf{Datapoints indicated at Epoch 0 are after one epoch of training. Models were not evaluated before the first epoch.}\label{fig:veghist}}
\end{figure}

\begin{figure}[h]
    \includegraphics[width=\textwidth]{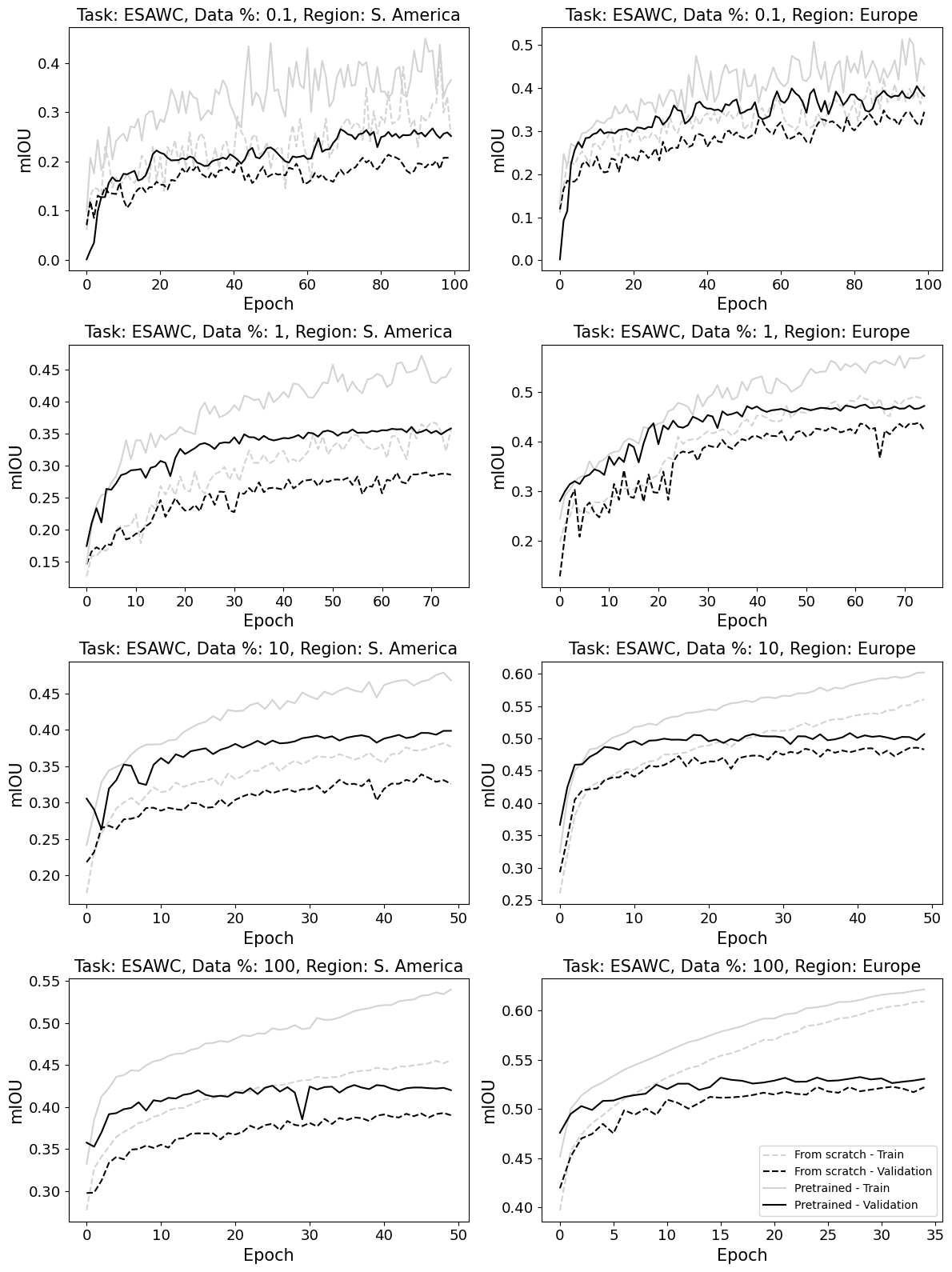}
  \caption{Training histories for ESA World Cover prediction. \textbf{Datapoints indicated at Epoch 0 are after one epoch of training. Models were not evaluated before the first epoch.}\label{fig:esawchist}}
\end{figure}

Training histories for MODISVEG and ESAWC can be seen in Figures~\ref{fig:veghist} and \ref{fig:esawchist} respectively. We did not tune the optimiser extensively, beyond changing the learning rate to achieve a reasonable rate of convergence. Note that the model was not evaluated before the first epoch (Epoch 0), so datapoints indicated at Epoch 0 are after one epoch of training.

\end{document}